\def\year{2016}
\documentclass[letterpaper]{article}
\usepackage{aaai16}
\usepackage{times}
\usepackage{helvet}
\usepackage{courier}
\usepackage{url}
\usepackage{latexsym}
\usepackage{amsmath}
\usepackage{amssymb}
\usepackage{amsthm}
\usepackage{graphicx}
\usepackage[tight,footnotesize]{subfigure}
\usepackage{multirow}
\usepackage{tabularx}
\usepackage{array}
\usepackage{algpseudocode}
\usepackage{algorithm}
\usepackage{algpseudocode}
\usepackage{float}
\usepackage{color}
\usepackage[justification=centering]{caption}
\usepackage{changepage}
\usepackage{wrapfig}
\begin{document}
	
	\title{Application of Statistical Relational Learning to Hybrid Recommendation Systems}

	\author{Shuo Yang* \and Mohammed Korayem \and Khalifeh AlJadda \and
	Trey Grainger \and Sriraam Natarajan* \\ 
	* School of Informatics and Computing, Indiana University Bloomington, IN 47408 \\
	CareerBuilder,Norcross, GA 30092 
	}

	\maketitle
	\begin{abstract}
	\begin{quote}
	 Recommendation systems usually involve exploiting the relations among known features and content that describe items (content-based filtering) or the overlap of similar users who interacted with or rated the target item (collaborative filtering). To combine these two filtering approaches, current model-based hybrid recommendation systems typically require extensive feature engineering to construct a user profile. Statistical Relational Learning (SRL) provides a straightforward way to combine the two approaches. However, due to the large scale of the data used in real world recommendation systems, little research exists on applying SRL models to hybrid recommendation systems, and essentially none of that research has been applied on real big-data-scale systems. In this paper, we proposed a way to adapt the state-of-the-art in SRL learning approaches to construct a real hybrid recommendation system. Furthermore, in order to satisfy a common requirement in recommendation systems (i.e. that false positives are more undesirable and therefore penalized more harshly than false negatives), our approach can also allow tuning the trade-off between the precision and recall of the system in a principled way. Our experimental results demonstrate the efficiency of our proposed approach as well as its improved performance on recommendation precision. 
	\end{quote}
	\end{abstract}

\section{Introduction}

\begin{figure*}[htbp]
	\begin{minipage}[b]{\textwidth}
		\vspace{-15pt}
		\centering
		\includegraphics[width=0.8\textwidth]{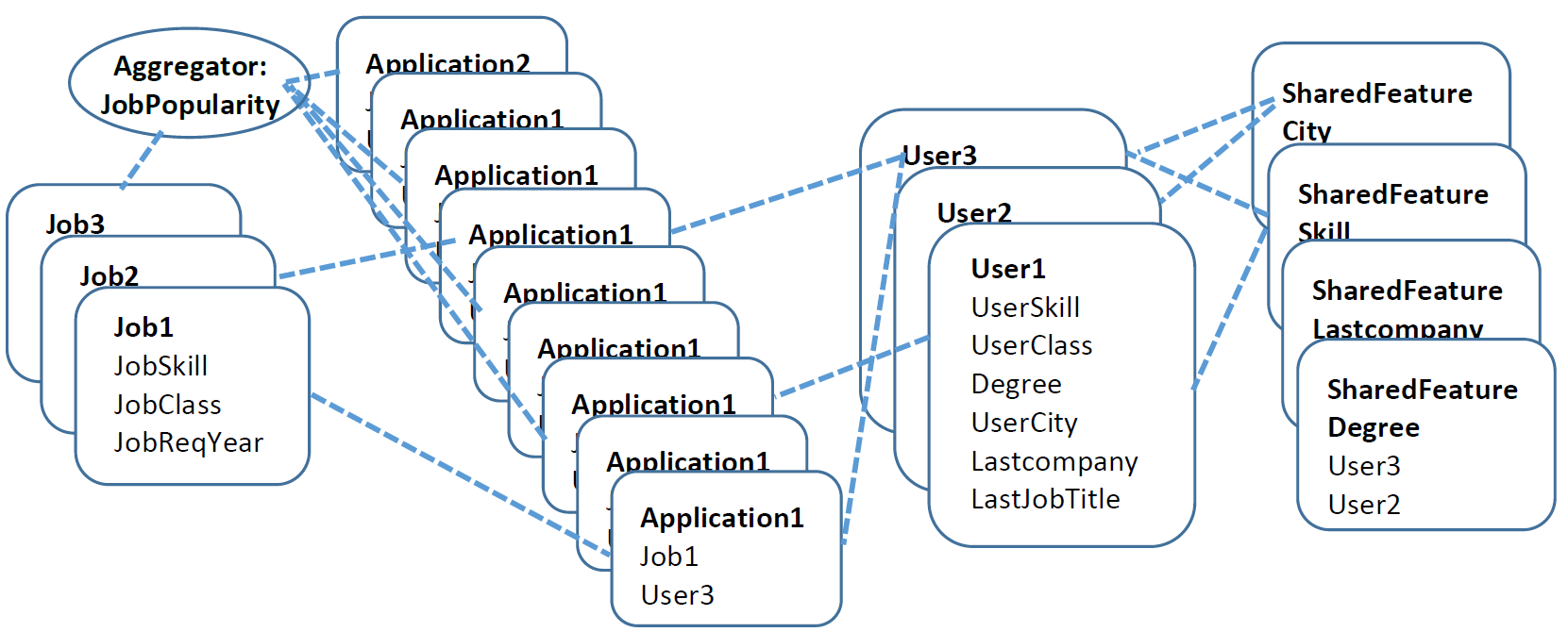}
	\end{minipage}
	\caption{ \small Job recommendation domain}
	\label{rgraph}
	\vspace{-2mm}
\end{figure*}

With their rise in prominence, recommendation systems have greatly alleviated information overload for their users by providing personalized suggestions for countless products such as music, movies, books, housing, jobs, and etc. We consider a specific recommender system domain, that of job recommendations, and propose to develop a novel method for this domain using Statistical Relational Learning. This domain easily scales to billions of items including user resumes and job postings, as well as even more data in the form of user interactions between these items. CareerBuilder, the source of the data for our experiments, operates one of the largest job boards in the world. It has millions of job postings, more than 60 million actively-searchable resumes, over one billion searchable documents, and receives several million searches per hour~\cite{aljadda2014crowdsourced}. 
The scale of the data is not the only interesting aspect of this domain, however. The job recommendations use case is inherently relational in nature, readily allowing for graph mining and relational learning algorithms to be employed. 
As Figure~\ref{rgraph} shows, very similar kinds of relationships exist among the jobs that are applied to by the same users and among the users who share similar preferences. 
   
One of the popular recommender approaches is {\em content-based filtering}~\cite{Basu98recommendationas}, which exploits the relations between (historically) applied-to jobs and similar features among new job opportunities for consideration (with features usually derived from textual information). An alternative recommendation approach is based on {\em collaborative filtering}~\cite{Breese1998}, which makes use of the fact that users who are interested in the same items generally have similar preferences for additional items. Clearly, using both types of information together can potentially yield a more powerful recommendation system, which is why model-based hybrid recommender systems were developed~\cite{Basilico2004}. While successful, these systems typically need extensive feature engineering to make the combination practical. 

Our hypothesis that we sought to verify empirically was that recent advancements in the fields of machine learning and Artificial Intelligence could lead to powerful and deployable recommender systems. In particular, we assessed leveraging Statistical Relational Learning (SRL)~\cite{srlbook}, which combines the representation abilities of rich formalisms such as first-order logic or relational logic with the ability of probability theory to model uncertainty. We employed a state-of-the-art SRL formalism for combining content-based filtering and collaborative filtering. SRL can directly represent the probabilistic dependencies among the attributes from different objects that are related with each other through certain connections (in our domain, for example, the jobs applied to by the same user or the users who share the same skills or employers). SRL models remove the necessity for an extensive feature engineering process, and they do not require learning separate recommendation models for each individual item or user cluster, a requirement for many standard model-based recommendation systems~\cite{Pazzani1997}.  

We propose a hybrid model combining content-based filtering and collaborative filtering that is learned by an efficient statistical relational learning approach - Relational Functional Gradient Boosting(RFGB)~\cite{Natarajan2012}. Specifically, we define the target relation as $Match(User, Job)$ which indicates that the user--job pair is a match when the grounded relation is true, hence that job should be recommended to the target user. The task is to predict the probability of this target relation $Match(User, Job)$ for users based on the information about the job postings, the user profile, the application history, as well as application histories of users that have the similar preferences or profiles as the target user. RFGB is a boosted model which contains multiple relational regression trees with additive regression values at the sink node of each path. Our hypothesis is that these trees can capture many of the weak relations that exist between the target user and the job with which he/she is matched.

In addition, this domain has practical requirements which must be considered. For example, we would rather overlook some of the candidate jobs that could match the users (false negatives) than send out numerous spam emails to the users with inappropriate job recommendations (false positives). The cost matrix thus does not contain uniform cost values, but instead needs to represent a higher cost for the user--job pairs that are false positives compared to those that are false negatives, i.e. precision is preferred over recall. To incorporate such domain knowledge within the cost matrix, we adapted the previous work~\cite{Yang14}, which extended RFGB by introducing a penalty term into the objective function of RFGB so that the trade-off between the precision and recall can be tuned during the learning process. 

In summary, we considered the problem of matching a user with a job and developed a hybrid content-based filtering and collaborative filtering approach. We adapted a successful SRL algorithm for learning features and weights and are the first to implement such a system in a real-world big data context. Our algorithm is capable of handling different costs for false positives and false negatives making it extremely attractive for deploying within many kinds of recommendation systems, including those within the domain upon which we tested.
\section{Related Work}


 Recommendation systems usually handle the task of estimating the relevancy or ratings of items for certain users based on information about the target user--item pair as well as other related items and users. The recommendation problem is usually formulated as $f: U\times I \rightarrow R$ where $U$ is the space of all users, $I$ is the space of all possible items and $f$ is the utility function that projects all combinations of user-item pairs to a set of predicted ratings $R$ which is composed by nonnegative integers. For a certain user $u$, the recommended item would be the item with the optimal utility value, i.e. $u_i^* = argMax_{u \in U}  f(u,i)$. The user space $U$ contains the information about all the users, such as their demographic characteristics, while the item space $I$ contains the feature information of all the items, such as the genre of the music, the director of a movie, or the author of a book.

Generally speaking, the goal of \textit{Content-based filtering} is to define recommendations based upon feature similarities between the items being considered and items which a user has previously rated as interesting~\cite{Adomavicius2005}, i.e. for the target user-item rating $f(\hat u, \hat i)$, \textit{Content-based filtering} would predict the optimal recommendation based on the utility functions of $f(\hat u, I_h)$ which is the historical rating information of user $\hat u$ on items($I_h$) similar with $\hat i$. 
Originated from information retrieval and information filtering, most content-based filtering systems are applied to items that are rich in textual information. From this textual information, item features $I$ are extracted and represented as \textit{keywords} with respective weighting measures calculated by certain mechanisms such as the \textit{term frequency/inverse document frequency (TF/IDF)} measure~\cite{Salton89}. The feature space of the user $U$ is then constructed from the feature spaces of items that were previously rated by that user through various keyword analysis techniques such as averaging approach~\cite{Rocchio71}, Bayesian classifier~\cite{Pazzani1997} and etc. Finally, the utility function of the target user-item pair $f(\hat u, \hat i)$ is calculated by some scoring heuristic such as the cosine similarity~\cite{Salton89} between the user profile vector and the item feature vector or some traditional machine learning models~\cite{Pazzani1997}. 

On the other hand, the goal of the \textit{collaborative filtering} is to recommend items by learning from users with similar preferences~\cite{Adomavicius2005,Su2009,NIPS2015_5938}, i.e. for the target user-item rating $f(\hat u, \hat i)$, \textit{Collaborative filtering} builds its belief in the best recommendation by learning from the utility functions of $f(U_s, \hat i)$ which is the rating information of the user set $U_s$ that has similar preferences as the target user $\hat u$. The commonly employed approaches fall into two categories: \textit{memory-based(or heuristic-based)} and \textit{model-based} systems. The heuristic-based approaches usually predict the ratings of the target user-item pair by aggregating the ratings of the most similar users for the same item with various aggregation functions such as mean, similarity weighted mean, adjusted similarity weighted mean(which uses relative rating scales instead of the absolute values to address the rating scale differences among users), etc. The set of most similar users and their corresponding weights can be decided by calculating the correlation (such as Pearson Correlation Coefficient~\cite{Resnick94}) or distance (such as cosine-based~\cite{Breese1998} or mean squared difference) between the rating vectors of the target user and the candidate user on common items. Whereas model-based algorithms are used to build a recommendation system by training certain machine learning models~\cite{Salakhutdinov2007,Breese1998,Si03,sahoo2010hidden} based on the ratings of users that belong to the same cluster or class as the target user. 
Hence, prior research has focused on applying statistical relational models to collaborative filtering systems~\cite{Getoor99,newton2004hierarchical,gao2007recommendation,huang2005unified}.

There are \textit{Hybrid approaches} which combine collaborative filtering and content-based filtering into a unified system ~\cite{de2010combining,balabanovic1997fab,Basilico2004}. For instance Basilico et al. ~\cite{Basilico2004} unified content-based and collaborative filtering by engineering the features based on various kernel functions, then trained a simple linear classifier (Perceptron) in this engineered feature space. 


The most related work to ours is \cite{HoxhaR13}, where they proposed to use Markov Logic Networks to build hybrid models combining content-based filtering and collaborative filtering. Their work only employed one type of probabilistic logic model, which is demonstrated later in this paper to not be the best one, and it did not consider the special requirement of many recommendation systems that precision should be preferred over recall (or at least that the relative weight of the two should be configurable).


\section{Building Hybrid Recommendation Systems with SRL Models}

In order to represent the data in a flat table, the standard model-based recommendation systems need an exhaustive feature engineering process to construct the user profile by aggregating the attributes over all the similar users who share the same background or similar preferences as the target user. The aggregation-based strategies are necessary because the standard algorithms require a regular flat table to represent the data. However, the number of similar users related to the target user may vary a lot among different individuals. For example, users with common preferences could have more similar users than the users with unique tastes. 

\begin{figure}[htbp]
	\centering
	\includegraphics[width=0.4\textwidth]{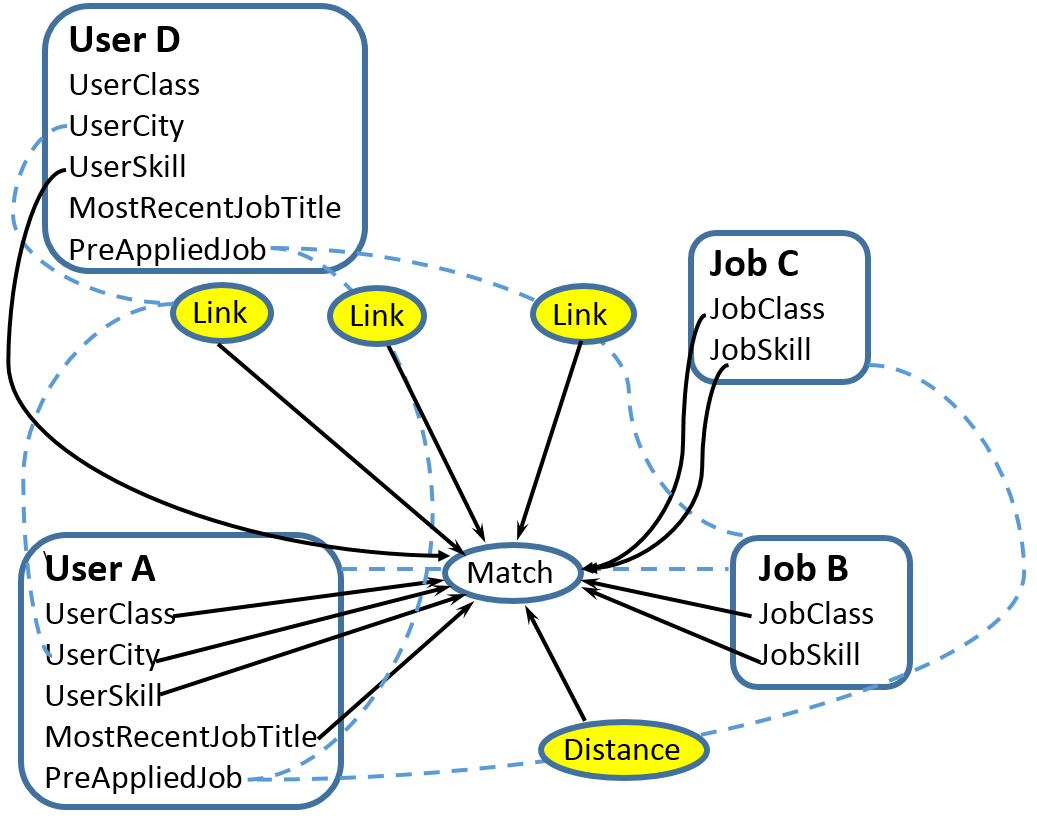}
	\caption{ \small Template Model of a Sample RDN Learned from Class22. The target is $Match(UserA, JobB)$ while the related objects are User D (introduced by the link nodes) and previous applied Job C. Note that D and C are first-order variables which could have multiple groundings for different target user--job pairs. }
	\label{rdn}
	\vspace{-2mm}
\end{figure}

We propose to employ SRL for the challenging task of implementing a hybrid recommendation system. Specifically, we consider the formulation of Relational Dependency Networks (RDN)~\cite{Neville07}, which are approximate graphical models that are inferred using the machinery of Gibbs sampling.
Figure~\ref{rdn} shows a template model of RDNs learned from our experiment. As can be seen, other than the attributes of the target user A and target job B, it also captures the dependencies between the target predicate $Match(A, B)$ and attributes from the similar user D and previous applied job C. 
As an approximation of Bayesian Networks, Dependency Networks (DNs) make the assumption that the joint distributions can be approximated as the product of the individual conditional probability distributions and that these conditional probability distributions are independent from each other. 
RDNs extend DNs to relational data and are considered as one of the most successful SRL models that have been applied to real-world problems. Hence, we propose to construct a hybrid recommendation system by learning an RDN using a state-of-the-art learning approach--Relational Functional Gradient Boosting(RFGB) which has been proven to be one of the most efficient relational learning approaches~\cite{Natarajan2012}. 

The following subsections will first introduce the basic concept of an RFGB, then cover the way we incorporate domain knowledge on the cost matrix so the proposed hybrid recommendation system can improve the confidence of recommended jobs. 

\subsection{Relational Functional Gradient Boosting}
\label{rfgb}
When fitting a probabilistic model $P(y|x)$, standard gradient ascent approaches start with initial parameters $\theta_0$ and iteratively add the gradient ($\Delta_i$) of an objective function with respect to $\theta_i$.  
Friedman \cite{friedman01} proposed an alternate approach where the objective function is represented using a regression function $\psi$ over the examples $\mathbf{x}$, and the gradients are performed w.r.t. $\psi(x)$. Similar to parametric gradient descent, after $n$ iterations of functional gradient descent, $\psi_n(x)$ $=$ $\psi_0(x) + \Delta_1(x)$ $+$ $\cdots$ $+$ $\Delta_n(x)$.

Each gradient term ($\Delta_m$) is a set of training examples and regression values given by the gradient w.r.t $\psi_m(x_i)$, i.e., $<x_i , \Delta_m(x_i) = \frac{\partial LL(\mathbf{x})}{\partial \psi_m(x_i)}>$. To generalize from these regression examples, a regression function $\hat{\psi}_m$ (generally regression tree)  is learned to fit to the gradients. The final model $\psi_m = \psi_0 + \hat{\psi}_1 + \cdots + \hat{\psi}_m$ is a sum over these regression trees. Functional gradient ascent is also known as functional gradient boosting (FGB) due to this sequential nature of learning models.

FGB has been applied to relational models~\cite{Natarajan2012,karwath08,PolicyGradient08,ijcaiImitation} due to its ability to learn the structure and parameters of these models simultaneously. Gradients are computed for every groundings/instantiation of the target first-order predicate. In our case, the grounding \textit{Match(John, Software Engineer)} of the target predicate \textit{Match(User, Job)} could be one example. Relational regression trees~\cite{tilde} are learned to fit the $\psi$ function over the relational regression examples. Since the regression function $\psi: X \rightarrow (-\infty, \infty)$ is unbounded, a sigmoid function over  $\psi$  is commonly used to represent conditional distributions. Thus the RFGB log-likelihood function is: $LL = $
\begin{equation*}
\hspace{-3mm}
\sum_i \log P(y_i=\hat{y}_i;\textbf{X}_i) = \sum_i \log \frac{1}{1 + \exp(-\hat{y}_i\cdot\psi(y_i=\hat{y}_i;\textbf{X}_i))}
\end{equation*}
where $y_i$ corresponds to a target grounding of example $i$  with parents $\mathbf{X}_i$. 
In our case, the target predicate is \textit{Match(User, Job)}, and the parents $\mathbf{X}_i$ would be the attributes of the target user and target job, and the jobs previously applied to by the target user and similar users sharing the same preferences. $\hat{y}_i$ is the true label for a user--job pair which is $1$ for a positive matching pair and $0$ for a negative matching pair. The key assumption is that the conditional probability of a target grounding $y_i$, given all the other predicates, is modeled as a sigmoid function. 

The gradient w.r.t. $\psi(y_i=\hat{y}_i;\textbf{X}_i)$ is  
\begin{equation}
\label{eq:grad}
\frac{\partial LL(\mathbf{x})}{\partial \psi(y_i=\hat{y}_i;\textbf{X}_i)} = I(\hat{y}_i= Match) - P(y_i=Match ;\textbf{X}_i)
\end{equation}
which is the difference between the true observation ($I$ is the indicator function) and the current predicted probability of the match being true. Note the indicator function, $I$  returns $1$ for positives and $0$ for negatives. Hence the positive gradient terms for positive examples push the regression values closer to $\infty$ and thereby the probabilities closer to 1, whereas for negative examples, the regression values are pushed closer to $-\infty$ and the probabilities closer to $0$. 

\subsection{Cost Sensitive Learning with RFGB}

Following the work of Yang et al.~\cite{Yang14}, we propose to construct a hybrid job recommendation system by learning a cost-sensitive RDN. 


As shown in equation~\ref{eq:grad}, the magnitude (absolute value) of the gradient in RFGB only depends on how well the current model fits the example. If it fits well, the probability of the positive example given the current model would be close to 1 (0 for negative examples), and the gradient that will be assigned to such examples as the training weights would approach 0. If it does not, the predicted probability of the example would be far from the true label and hence cause the boosting algorithm to attach a high weight to that example. As a result, this method treats both false positive and false negative examples in the same way. Since most of the relational data suffers from class imbalance, where negative instances are much higher cost than positive instances, the negative outliers would easily dominate the classification boundary after a few iterations.  
So, Yang et al.~\cite{Yang14} proposed a cost-sensitive relational learning approach which is able to address these issues and model the target task more faithfully. This is achieved by adding a term to the objective function that penalizes false positives and false negatives differently.
They defined the cost function as:
\begin{equation*}
c(\hat{y}_i, y_i)= \alpha \, I(\hat{y}_i=1\wedge y_i=0) \, + \, \beta I(\hat{y}_i=0 \wedge y_i=1),
\end{equation*}
where $\hat{y}_i$ is the true label of the $i^{th}$ instance and $y_i$ is the predicted label. $I(\hat{y}_i=1 \wedge y_i=0)$ is $1$ for false negatives (in our case, the matching user--job pair that is predicted as mis-matching) and $I(\hat{y}_i=0 \wedge y_i=1)$ is $1$ for false positives (in our case, the mis-matching user--job pair that is classified as matching). This cost function was hence being introduced into the normalization term of the objective function as:
\begin{equation*}
\log J=  \sum_i  \psi(y_i;\textbf{X}_i)-\log\sum_{y^\prime_i}\exp
\left\{ \psi(y^\prime_i;\textbf{X}_i) + c(\hat{y}_i,y^\prime_i) \right\}
\end{equation*}

Thus, in addition to simple log-likelihood of the examples, the algorithm also takes into account these additional costs.

Then, the gradient of the objective function w.r.t $\psi(y_i=1;\textbf{X}_i)$ can be calculated by:
\begin{equation}
\Delta = I(\hat{y}_i=Match) \, - \, \lambda P(y_i=Match;\textbf{X}_i).
\label{simp_grad}
\end{equation}
where $\lambda\, =$
\begin{equation}
\begin{array}{l}
\left\{ \begin{array}{ll}
\displaystyle{For \ \ matching \ \ User-Job \ \ pairs:} \\
\displaystyle{\frac{1}{P(y^\prime=Match;\textbf{X}_i)+P(y^\prime=MisMatch;\textbf{X}_i)\cdot e^{\alpha}}} \\ 
\\
\displaystyle{For\ \ mismatching\ \ User-Job\ \ pairs:} \\
\displaystyle{\frac{ e^{\beta}}{P(y^\prime=Match;\textbf{X}_i)\cdot e^{\beta} +P(y^\prime=MisMatch;\textbf{X}_i)}} 
\end{array} \right.
\end{array}
\label{lambda}
\end{equation}

As shown above, the cost function $c(\hat{y}_i, y_i)$ is controlled by $\alpha$ when a positive example is misclassified, while being controlled by $\beta$ when a negative example is misclassified.
 
\textit{Generally, if $\alpha <0$ ($\beta < 0$), the algorithm is more tolerant of misclassified positive (negative) examples. Alternately, if $\alpha > 0$ ($\beta > 0$), the algorithm penalizes misclassified positive (negative) examples even more than standard RFGB.} Thus, the influence of positive and negative examples on the final learned distribution can be directly controlled by tuning the parameters $\alpha$ and $\beta$.

Now, consider the special requirement on the cost matrix in most job recommendation systems that we would rather miss certain candidate jobs which to some extent match the target user than send out recommendations that are not appropriate to the target user. In other word, we prefer high precision as long as the recall maintains above such a reasonable value that the system would not return zero recommendations for the target user. 

Since $\alpha$ is the parameter controlling the weights of false negative examples, we simply assign it as 0 which makes $\lambda=1 \, / \, \sum_{y^\prime} \, [P(y^\prime; \textbf{X}_i)] = 1$ for misclassified positive examples. As a result, the gradient of the positive examples is the same as it was in the original RFGB settings. 

For the false positive examples, we use a harsher penalty on them, so the algorithm would put more effort into classifying them correctly in the next iteration. 
According to Equation~\ref{lambda}, when it is a negative example ($\hat{y}_i = 0$), we have
\begin{equation}
\lambda= \frac{1}{P(y^\prime=Match;\textbf{X}_i)+P(y^\prime=MisMatch;\textbf{X}_i)\cdot e^{-\beta}}. \nonumber
\end{equation}
As $\beta \rightarrow \infty$, $e^{-\beta} \rightarrow 0$, hence $\lambda \rightarrow 1\,/\,P(y_i=Match; \textbf{X}_i)$, so
\begin{equation*}
\Delta = 0 \, - \, \lambda P(y_i=Match;\textbf{X}_i) \rightarrow -1 
\end{equation*}

This means the gradient is pushed closer to its maximum magnitude $|-1|$, no matter how close the predicted probability is to the true label. On the other hand, when $\beta \rightarrow -\infty$, then $\lambda \rightarrow  0$, hence $\Delta \rightarrow 0$, which means that the gradients are pushed closer to their minimum value of 0. So, in our experiment, we set $\beta > 0$, which amounts to putting a large negative weight on the false positive examples.




Consider a medical diagnosis task, where we would wish to correctly classify as many positive examples as possible, while at the same time, avoid over-fitting the negative examples.  In such a case, setting $\beta < 0$ can satisfy the domain requirements on the cost matrix (i.e. classifying negative example as positive is to some extent tolerable), as well as handle special properties of the data (i.e. that the class is highly imbalanced with negative examples as the majority) at the same time. 

In job recommendation system, by contrast, the major goal is typically not to have mis-classified false positive examples. As a result, we need to eliminate the noise/outliers in negative examples as much as possible. Most algorithms generate negative examples by randomly drawing objects from two related variables, and the pairs that are not known as positively-related for the given facts are assumed to be a negative pair. However, in our case, if we randomly draw instances from $User$ and $Job$, and assume it is a negative example if that grounded user never applied to that grounded job, it could introduce a lot of noise into the data since not applying could be the result of any number of reasons. For example, it could simply be due to the job never being seen by the user. Hence, instead of generating negative instances following a ``closed-world assumption'', as most of the relational data did, we instead generated the negative examples by extracting the jobs that were sent to the user as recommendations but were not applied to by the user. In this way, we can guarantee that this User--Job pair is indeed not matching. 

\begin{table*}[!htbp]
	\caption{Domains} 
	\vspace{-3mm}
	\centering
	\begin{tabular}{|c|c|c|c|c|c|c|c|}
		\hline
		\multirow{2}{*}{~} & \multirow{2}{*}{JobTitle} & \multicolumn{3}{c}{Training} & \multicolumn{3}{|c|}{Test} \\ \cline{3-8}
		~ & ~& pos & neg & facts & pos & neg & facts \\ \hline
		Class20 & Retail Sales Consultant & 224 & 6973 & 13340875 & 53 & 1055 & 8786938 \\ \hline
		
	\end{tabular}	
	\label{domains}
\end{table*}
\begin{table*}[!htb]
	\caption{Results }
	\vspace{-3mm}	
	\centering
	\begin{tabular}{|c|c|c|c|c|c|c|c|}
		\hline
		\multicolumn{2}{|c|}{~} & \textbf{FPR} & FNR & Precision & Recall & Accuracy & AUC-ROC \\ \hline
		
		~ & Content-based Filtering & 0.537 & 0.321 & 0.060 & 0.679 & 0.473 & 0.628  \\ \cline{2-8}
		~ & Soft Content-based Filtering & \textbf{0.040} & 0.868  & \textbf{0.143} & 0.132 & \textbf{0.921} & 0.649  \\ \cline{2-8}
		Class20 & Hybrid Recommender & 0.516 & \textbf{0}  & 0.089 & \textbf{1.0} & 0.509 & \textbf{0.776}   \\ \cline{2-8}
		~ & Soft Hybrid Recommender & 0.045  & 0.906 & 0.096 & 0.094 & 0.914 & 0.755  \\ \hline

	\end{tabular}
	\label{results}
\end{table*}

\section{Experiments}
\label{exp}
\vspace{2mm}

We extracted 4 months of user job application history and active job posting records and evaluated our proposed model on that data. Our intention was to investigate whether our proposed model can efficiently construct a hybrid recommendation system with cost-sensitive requirements by explicitly addressing the following questions:
\begin{description}	
	\item{\bf(Q1)} How does combining collaborative filtering improve the performance compared with content-based filtering alone?
	\item{\bf(Q2)} Can the proposed cost-sensitive SRL learning approach reduce false positive prediction without sacrificing too much on the other evaluation measurements?
\end{description}

To answer these questions, we extracted 9 attributes from user resumes as well as job postings, which are defined as first-order predicates: \textit{JobSkill(jobid, skillid)}, \textit{UserSkill(userid, skillid)}, \textit{JobClass (jobid, classid)}, \textit{UserClass(userid, classid)}, \textit{PrAppliedJob(userid, jobid)}, \textit{UserJobDis(userid, jobid, distance)},  \textit{UserCity(userid, cityname)}, \textit{MostRecentCompany(userid, companyid)}, \textit{mostRecentJobTitle(userid, jobtitle)}. 

There are 707820 total job postings in our sample set, and the number of possible instances the first order variables can take is shown below. 
\begin{table}[htbp]
	\vspace{-2mm}
	\centering
	\small
	\begin{tabular}{|c|c|c|c|}
		\hline
		Variable Name & skillDid & classDid & distance  \\ \hline
		Num of Instances & 8534 & 1867 & 4  \\ \hline
		Variable Name & cityname & companyid & jobtitle \\ \hline
		Num of Instances & 22137 & 1154623 & 823733 \\ \hline
	\end{tabular} 
	\label{variables}	
\end{table}

Information on the \textit{JobClass} and \textit{UserClass} are extracted based upon the work of Javed et al.~\cite{Javed2015}.
The other features related to users are \textit{UserSkill}, \textit{UserCity}, \textit{MostRecentCompany} and \textit{mostRecentJobTitle} which are either extracted from the user's resume or the user's profile document, whereas the job feature \textit{JobSkill} represents a desired skill extracted from the job posting. Predicate \textit{UserJobDis} indicates the distance between the user(first argument) and the job(second argument), which is calculated based on the user and job locations extracted from respective documents. The \textit{UserJobDis} feature is discretized into 4 classes (1: $<15$ mile; 2: [15 miles, 30 miles); 3: [30 miles, 60 miles]; 4: $>60$ miles). The predicate \textit{PrAppliedJob} defines the previous applied jobs and serves as both an independent predicate which indicates whether the target user is in a cold start scenario, as well as acting as a bridge which introduces into the searching space the attributes of other jobs related to the target user during the learning process. 

We also use three additional first-order predicates: \textit{CommSkill(userid1, userid2)}, \textit{CommClass (userid1, userid2)} and \textit{CommCity(userid1, userid2)} which are induced from the given groundings of the predicates \textit{UserSkill}, \textit{UserClass} and \textit{UserCity} and also serve as bridges which introduce features of other users who share the similar background with the target user.

The performance of our model is evaluated in 1 user classes, each of which has its data scale description shown in Table~\ref{domains}.  

For each of these user classes, we experimented with our proposed model using first-order predicates of the content-based filtering alone, as well the first-order predicates of both content-based filtering and collaborative filtering. 

As Table~\ref{results} shows, although the two approaches show similar performance on False Positive Rate, Precision, and Accuracy, the hybrid recommendation system improves a lot on the False Negative Rate, Recall and AUC-ROC compared with content-based filtering alone, especially on the Recall (reached 1.0 for all three of the user classes). So, question {\bf(Q1)} can be answered affirmatively. The hybrid recommendation system improves upon the performance of content-based filtering alone, by taking into consideration the information of similar users who have the same expertise or location as the target user. 

The first column of Table~\ref{results} shows the False Positive Rate which we want to reduce. As the numbers shown, the soft margin approach greatly decreases the FPR compared with prior research which does not consider the domain preferences on the cost matrix. It also significantly improves the accuracy at the same time. Note that, although it seems that recall has been considerably sacrificed, our goal here is not to capture all the matching jobs for the target user, but instead to increase the confidence on the recommendations we are giving to our users. Since we may have hundreds of millions of candidates and jobs in the data pool, we can usually guarantee that we will have a sufficient number of recommendations even with relatively low recall. Hence, question {\bf(Q2)} can be also be answered affirmatively. Moreover, our proposed system can satisfy various requirements on the trade-off of precision and recall for different practical consideration by tuning the parameters $alpha$ and $beta$. If one does not want the recall too low, in order to guarantee the quantity of recommendations, one can simply decrease the value of $beta$; if one does not want the precision too low, in order to improve the customer satisfaction, one can just increase the value of $beta$. 


It is worth mentioning that we also tried to experiment with Markov Logic Networks on the same data by using \textit{Alchemy2} \cite{Alchemy09}. However, it failed after continuously running for three months due to the large scale of our data. This underscores one of the major contributions of this research in applying SRL using a hybrid approach in a real-world large-scale job recommendation system.
\section{Conclusion}

We proposed an efficient statistical relational learning approach to construct a hybrid job recommendation system which can also satisfy the unique cost requirements regarding precision and recall of a specific domain. The experiment results show the ability of our model to reduce the rate of inappropriate job recommendations. Our contribution includes: i. we are the first to apply statistical relational learning models to a real-world large-scale job recommendation system; ii. our proposed model not only proves to be the most efficient SRL learning approach, but also demonstrates its ability to further reduce false positive predictions; iii. the experiment results reveal a promising direction for future hybrid recommendation systems-- with proper utilization of first-order predicates, an SRL-model-based hybrid recommendation system can not only prevent the necessity for exhaustive feature engineering or pre-clustering, but can also provide a robust way to solve the cold-start problem.


\bibliographystyle{aaai}
\bibliography{bib}

\begin{thebibliography}{}

\bibitem[\protect\citeauthoryear{Adomavicius and
  Tuzhilin}{2005}]{Adomavicius2005}
Adomavicius, G., and Tuzhilin, A.
\newblock 2005.
\newblock Toward the next generation of recommender systems: A survey of the
  state-of-the-art and possible extensions.
\newblock {\em IEEE Trans. on Knowl. and Data Eng.} 17(6):734--749.

\bibitem[\protect\citeauthoryear{AlJadda \bgroup et al\mbox.\egroup
  }{2014}]{aljadda2014crowdsourced}
AlJadda, K.; Korayem, M.; Grainger, T.; and Russell, C.
\newblock 2014.
\newblock Crowd sourced query augmentation through semantic discovery of
  domain-specific jargon.
\newblock In {\em 2014 IEEE International Conference on Big Data},  808--815.
\newblock IEEE.

\bibitem[\protect\citeauthoryear{Balabanovi{\'c} and
  Shoham}{1997}]{balabanovic1997fab}
Balabanovi{\'c}, M., and Shoham, Y.
\newblock 1997.
\newblock Fab: content-based, collaborative recommendation.
\newblock {\em Communications of the ACM} 40(3):66--72.

\bibitem[\protect\citeauthoryear{Basilico and Hofmann}{2004}]{Basilico2004}
Basilico, J., and Hofmann, T.
\newblock 2004.
\newblock Unifying collaborative and content-based filtering.
\newblock In {\em Proceedings of the Twenty-first International Conference on
  Machine Learning}.

\bibitem[\protect\citeauthoryear{Basu, Hirsh, and
  Cohen}{1998}]{Basu98recommendationas}
Basu, C.; Hirsh, H.; and Cohen, W.
\newblock 1998.
\newblock Recommendation as classification: Using social and content-based
  information in recommendation.
\newblock In {\em Fifteenth National Conference on Artificial Intelligence},
  714--720.
\newblock AAAI Press.

\bibitem[\protect\citeauthoryear{Blockeel and Raedt}{1998}]{tilde}
Blockeel, H., and Raedt, L.~D.
\newblock 1998.
\newblock Top-down induction of first-order logical decision trees.
\newblock {\em Artificial Intelligence} 101:285--297.

\bibitem[\protect\citeauthoryear{Breese, Heckerman, and
  Kadie}{1998}]{Breese1998}
Breese, J.~S.; Heckerman, D.; and Kadie, C.
\newblock 1998.
\newblock Empirical analysis of predictive algorithms for collaborative
  filtering.
\newblock In {\em Proceedings of the Fourteenth Conference on Uncertainty in
  Artificial Intelligence},  43--52.
\newblock Morgan Kaufmann Publishers Inc.

\bibitem[\protect\citeauthoryear{De~Campos \bgroup et al\mbox.\egroup
  }{2010}]{de2010combining}
De~Campos, L.~M.; Fern{\'a}ndez-Luna, J.~M.; Huete, J.~F.; and Rueda-Morales,
  M.~A.
\newblock 2010.
\newblock Combining content-based and collaborative recommendations: A hybrid
  approach based on bayesian networks.
\newblock {\em International Journal of Approximate Reasoning} 51(7):785--799.

\bibitem[\protect\citeauthoryear{Friedman}{2001}]{friedman01}
Friedman, J.~H.
\newblock 2001.
\newblock Greedy function approximation: A gradient boosting machine.
\newblock {\em Annals of Statistics}  1189--1232.

\bibitem[\protect\citeauthoryear{Gao \bgroup et al\mbox.\egroup
  }{2007}]{gao2007recommendation}
Gao, Y.; Qi, H.; Liu, J.; and Liu, D.
\newblock 2007.
\newblock A recommendation algorithm combining user grade-based collaborative
  filtering and probabilistic relational models.
\newblock In {\em Fourth International Conference on Fuzzy Systems and
  Knowledge Discovery}, volume~1,  67--71.
\newblock IEEE.

\bibitem[\protect\citeauthoryear{Getoor and Sahami}{1999}]{Getoor99}
Getoor, L., and Sahami, M.
\newblock 1999.
\newblock Using probabilistic relational models for collaborative filtering.
\newblock In {\em Workshop on Web Usage Analysis and User Profiling
  (WEBKDD'99)}.

\bibitem[\protect\citeauthoryear{Getoor and Taskar}{2007}]{srlbook}
Getoor, L., and Taskar, B.
\newblock 2007.
\newblock {\em Introduction to Statistical Relational Learning}.
\newblock Adaptive computation and machine learning. MIT Press.

\bibitem[\protect\citeauthoryear{Hoxha and Rettinger}{2013}]{HoxhaR13}
Hoxha, J., and Rettinger, A.
\newblock 2013.
\newblock First-order probabilistic model for hybrid recommendations.
\newblock In {\em 12th International Conference on Machine Learning and
  Applications, {ICMLA} 2013},  133--139.

\bibitem[\protect\citeauthoryear{Huang, Zeng, and
  Chen}{2005}]{huang2005unified}
Huang, Z.; Zeng, D.~D.; and Chen, H.
\newblock 2005.
\newblock A unified recommendation framework based on probabilistic relational
  models.
\newblock {\em Available at SSRN 906513}.

\bibitem[\protect\citeauthoryear{Javed \bgroup et al\mbox.\egroup
  }{2015}]{Javed2015}
Javed, F.; Luo, Q.; McNair, M.; Jacob, F.; Zhao, M.; and Kang, T.~S.
\newblock 2015.
\newblock Carotene: A job title classification system for the online
  recruitment domain.
\newblock In {\em IEEE First International Conference on Big Data Computing
  Service and Applications},  286--293.

\bibitem[\protect\citeauthoryear{Karwath, Kersting, and
  Landwehr}{2008}]{karwath08}
Karwath, A.; Kersting, K.; and Landwehr, N.
\newblock 2008.
\newblock Boosting relational sequence alignments.
\newblock In {\em ICDM}.

\bibitem[\protect\citeauthoryear{Kok \bgroup et al\mbox.\egroup
  }{2009}]{Alchemy09}
Kok, S.; Sumner, M.; Richardson, M.; Singla, P.; Poon, H.; Lowd, D.; Wang, J.;
  and Domingos, P.
\newblock 2009.
\newblock The alchemy system for statistical relational {AI}.
\newblock Technical report, Department of Computer Science and Engineering,
  University of Washington, Seattle, WA.

\bibitem[\protect\citeauthoryear{Natarajan \bgroup et al\mbox.\egroup
  }{2011}]{ijcaiImitation}
Natarajan, S.; Joshi, S.; Tadepalli, P.; Kristian, K.; and Shavlik, J.
\newblock 2011.
\newblock Imitation learning in relational domains: A functional-gradient
  boosting approach.
\newblock In {\em IJCAI}.

\bibitem[\protect\citeauthoryear{Natarajan \bgroup et al\mbox.\egroup
  }{2012}]{Natarajan2012}
Natarajan, S.; Khot, T.; Kersting, K.; Gutmann, B.; and Shavlik, J.
\newblock 2012.
\newblock Gradient-based boosting for statistical relational learning: The
  relational dependency network case.
\newblock {\em Mach. Learn.} 86(1):25--56.

\bibitem[\protect\citeauthoryear{Neville and Jensen}{2007}]{Neville07}
Neville, J., and Jensen, D.
\newblock 2007.
\newblock {Relational Dependency Networks}.
\newblock {\em J. Mach. Learn. Res.} 8:653--692.

\bibitem[\protect\citeauthoryear{Newton and
  Greiner}{2004}]{newton2004hierarchical}
Newton, J., and Greiner, R.
\newblock 2004.
\newblock Hierarchical probabilistic relational models for collaborative
  filtering.
\newblock In {\em Workshop on Statistical Relational Learning, 21st
  International Conference on Machine Learning}.

\bibitem[\protect\citeauthoryear{Pazzani and Billsus}{1997}]{Pazzani1997}
Pazzani, M., and Billsus, D.
\newblock 1997.
\newblock Learning and revising user profiles: The identification ofinteresting
  web sites.
\newblock {\em Mach. Learn.} 27(3):313--331.

\bibitem[\protect\citeauthoryear{Rao \bgroup et al\mbox.\egroup
  }{2015}]{NIPS2015_5938}
Rao, N.; Yu, H.-F.; Ravikumar, P.~K.; and Dhillon, I.~S.
\newblock 2015.
\newblock Collaborative filtering with graph information: Consistency and
  scalable methods.
\newblock In {\em Advances in Neural Information Processing Systems 28}. Curran
  Associates, Inc.
\newblock  2107--2115.

\bibitem[\protect\citeauthoryear{Resnick \bgroup et al\mbox.\egroup
  }{1994}]{Resnick94}
Resnick, P.; Iacovou, N.; Suchak, M.; Bergstrom, P.; and Riedl, J.
\newblock 1994.
\newblock Grouplens: An open architecture for collaborative filtering of
  netnews.
\newblock  175--186.
\newblock ACM Press.

\bibitem[\protect\citeauthoryear{Rocchio}{1971}]{Rocchio71}
Rocchio, J.
\newblock 1971.
\newblock Relevance feedback in information retrieval.
\newblock In {\em The {SMART} {R}etrieval {S}ystem: {E}xperiments in
  {A}utomatic {D}ocument {P}rocessing}. Prentice-Hall Inc.
\newblock chapter~14,  313--323.

\bibitem[\protect\citeauthoryear{Sahoo, Singh, and
  Mukhopadhyay}{2010}]{sahoo2010hidden}
Sahoo, N.; Singh, P.~V.; and Mukhopadhyay, T.
\newblock 2010.
\newblock A hidden markov model for collaborative filtering.
\newblock {\em Management Information Systems Quarterly, Forthcoming}.

\bibitem[\protect\citeauthoryear{Salakhutdinov, Mnih, and
  Hinton}{2007}]{Salakhutdinov2007}
Salakhutdinov, R.; Mnih, A.; and Hinton, G.
\newblock 2007.
\newblock Restricted boltzmann machines for collaborative filtering.
\newblock In {\em Proceedings of the 24th International Conference on Machine
  Learning},  791--798.
\newblock ACM.

\bibitem[\protect\citeauthoryear{Salton}{1989}]{Salton89}
Salton, G.
\newblock 1989.
\newblock {\em Automatic Text Processing: The Transformation, Analysis, and
  Retrieval of Information by Computer}.
\newblock Boston, MA, USA: Addison-Wesley Longman Publishing Co., Inc.

\bibitem[\protect\citeauthoryear{Si and Jin}{2003}]{Si03}
Si, L., and Jin, R.
\newblock 2003.
\newblock Flexible mixture model for collaborative filtering.
\newblock In {\em ICML},  704--711.
\newblock AAAI Press.

\bibitem[\protect\citeauthoryear{Su and Khoshgoftaar}{2009}]{Su2009}
Su, X., and Khoshgoftaar, T.~M.
\newblock 2009.
\newblock A survey of collaborative filtering techniques.
\newblock {\em Adv. in Artif. Intell.} 2009:4:2--4:2.

\bibitem[\protect\citeauthoryear{Sutton \bgroup et al\mbox.\egroup
  }{2000}]{PolicyGradient08}
Sutton, R.; McAllester, D.; Singh, S.; and Mansour, Y.
\newblock 2000.
\newblock Policy gradient methods for reinforcement learning with function
  approximation.
\newblock In {\em NIPS}.

\bibitem[\protect\citeauthoryear{Yang \bgroup et al\mbox.\egroup
  }{2014}]{Yang14}
Yang, S.; Khot, T.; Kersting, K.; Kunapuli, G.; Hauser, K.; and Natarajan, S.
\newblock 2014.
\newblock Learning from imbalanced data in relational domains: {A} soft margin
  approach.
\newblock In {\em 2014 {IEEE} International Conference on Data Mining, {ICDM}
  2014},  1085--1090.

\end{thebibliography}

\end{document}